\documentclass{article}

\usepackage{PRIMEarxiv}

\usepackage[utf8]{inputenc} 
\usepackage[T1]{fontenc}    
\usepackage{hyperref}       
\usepackage{url}            
\usepackage{booktabs}       
\usepackage{amsfonts}       
\usepackage{nicefrac}       
\usepackage{microtype}      
\usepackage{lipsum}
\usepackage{fancyhdr}       
\usepackage{graphicx}       
\usepackage[round,authoryear]{natbib} 
\graphicspath{{media/}}     

\title{The study of short texts in digital politics: Document aggregation for topic modeling
}

\author{
  Nitheesha Nakka \\
  Pennsylvania State University \\
  University Park, PA\\
  \texttt{nvn5240@psu.edu} \\
   \And
  Omer F. Yalcin \\
  University of Massachusetts Amherst \\
  Amherst, MA\\
  \texttt{oyalcin@umass.edu} \\
  \And
  Bruce A. Desmarais\\
  Pennsylvania State University \\
  University Park, PA\\
  \texttt{bdesmarais@psu.edu} \\
  \And
  Sarah Rajtmajer\\
  Pennsylvania State University \\
  University Park, PA\\
  \texttt{smr48@psu.edu} \\
  \And
  Burt Monroe\\
  Pennsylvania State University \\
  University Park, PA\\
  \texttt{burtmonroe@psu.edu} \\
}

\begin{document}
\maketitle

\begin{abstract}
Statistical topic modeling is widely used in political science to study text. Researchers examine documents of varying lengths, from tweets to speeches. There is ongoing debate on how document length affects the interpretability of topic models. We investigate the effects of aggregating short documents into larger ones based on natural units that partition the corpus. In our study, we analyze one million tweets by U.S. state legislators from April 2016 to September 2020. We find that for documents aggregated at the account level, topics are more associated with individual states than when using individual tweets. This finding is replicated with Wikipedia pages aggregated by birth cities, showing how document definitions can impact topic modeling results.
\end{abstract}


\section{Topic Models and Short Text}
Political scientists regularly study corpora of short texts. For example, Twitter/x 
is an important platform for public engagement by politicians, and consequently an important source of text data for political scientists \citep[e.g.,][]{payson2022using, kim2021attention, tai2023official,butler2023male}. Political scientists have extensively used text analysis \citep{genovese2019politics, roberts2016introduction, mertens2019tweet,cirone2023asymmetric,gervais2020tweeting,cassell2021following} to understand corpora generated by politicians, 
and in particular topic models \citep{ grimmer2010bayesian,russell2022constituent}, for quantitative and qualitative study of Twitter data. 
But, despite the prolific use of topic modeling, the aggregation or disaggregation choices that define what constitutes a single document and document length is often overlooked.
Topic model output can vary greatly depending on document length \citep{hong2010empirical}. Therefore, to ensure robust and interpretable results, it is important to explore alternative approaches to defining documents. 

Furthermore, Twitter data is short text \citep{hong2010empirical}. 
In order to perform topic modeling on a corpus of tweets, researchers must decide whether and how to aggregate individual tweets into documents. Level of document aggregation can be considered a preprocessing step, as characterized by \citet{denny2018text}. This decision is meaningful, as prior work has shown that topic models are  
sensitive to document lengths and definitions \citep{hong2010empirical}. 
Our research objective is to explore how topic model results change when short documents are aggregated. While aggregating documents is a possibility with short text corpora studied across many fields, it is a particularly prevalent option in the textual data of interest to political scientists. Many of the corpora studied in political science arise in the context of hierarchically organized institutions (e.g., legislatures \citep{kim2021attention} nested in states,  judicial opinions nested in states \citep{goelzhauser2014judicial}, geographically partitioned news media \citep{ban2019newspapers}), that naturally define hierarchies with multiple levels.  We analyze a data set of one million randomly sampled tweets from U.S. state legislators \citep{kim2021attention}, posted between April 1st 2016 and September 30th 2020. The data represents the accounts of 4,084 total legislators with representation from all 50 states.\footnote{In 2017 Twitter changed their character limit from 140 to 280 and, approximately 25\% of the tweets in this sample were created before 2017. The character limit changes do not affect the comparison between aggregated vs dis-aggregated documents, especially since the majority of the tweets in our sample were created after 2017. Also the Wikipedia article analysis, that did not undergo any structural changes in article length, bolsters our findings.}

Given their recent popularity, with over 1,500 citations to the article introducing the R software \citep{roberts2019stm} in the last three years, we use structural topic models (STM)--- a probabilistic approach to topic modeling. 
We estimate four models: two in which the document is defined at the per-legislator level as an aggregation of a legislator's tweets during the time of study---one document for each individual legislator (i.e., 4,084 documents); 
and two in which the document is defined as the tweet (i.e., one million documents). 
The two models implemented for each document definition include a 60 and 120 topics.
In the online appendix, we implement a transformer-based approach and estimate four identical models using BERTopic.
Aggregating tweets to the legislator level is what \citet{hong2010empirical} refer to as the ``USER scheme'' in fitting topic models to tweets. They explain two potential advantages of the USER scheme. First, the longer documents created by aggregating tweets improves the ability of terms to discriminate among different topics. Second, as seen in Table~\ref{tab:time}, the model can be computed faster since the number of documents is much smaller. We find that the results differ significantly by document definition, with far more state-related topics 
generated using the 
per-legislator document definition in the STM approach
and BERTopic approach. 
Therefore, aggregating and/or dis-aggregating text can be considered an important part of the text pre-processing pipeline. 

To 
understand generalizability of our findings beyond the corpus of legislators' tweets, we conduct a similar analysis of U.S. persons' Wikipedia pages, aggregating by city of birth, which we present in the online appendix. In the analysis of Wikipedia pages, we find that aggregating pages based on city of birth results in a higher prevalence of state-related topics, which replicates our results on the Twitter corpus. This secondary application is designed primarily as a test of whether the patterns induced through aggregation of the Tweets are replicated in another corpus, and is less practical than aggregating tweets. Our results point to document aggregation as a particularly salient preprocessing step to consider in the study of state politics, as our applications indicate that aggregation can highlight threads of content that connect closely with the state level of variation.

\begin{table}[hbt!]
\centering
\caption{STM Compute Time in Hours: Tweets Document Definition vs. Legislator Document Definition.}
\label{tab:time}
\begin{tabular}{lll}
\toprule
&Tweets &Legislator \\
\midrule
\texttt{60-topics} & 10h & 1h \\
\texttt{120-topics} & 76h 30min & 2h \\
\bottomrule
\end{tabular}
\end{table}

\section{Structural Topic Model Results}
We 
follow the same preprocessing of document text and model summaries for all four models. Using the R package quanteda  \citep{benoit2018quanteda}, we removed any terms that did not appear in at least three documents. We also removed English stop words. During the preprocessing five legislators and their tweets are filtered out from the sample. To assure that our findings are robust to other model settings, for each document definition, we run one STM with 120 topics and one with 60 topics. In the online appendix we present measures of topic exclusivity and semantic coherence following the recommendations of \citet{roberts2014structural} with models that include 30, 60, 90, 120, and 150 topics. We find that both 60 and 120 topics provide the best balance between average exclusivity and average semantic coherence scores. To summarize the results, for each model we present the top ten terms in all topics, as is commonly seen in topic model literature \citep{grimmer2010bayesian,greene2014many,sharma2021fifty,tonidandel2022using}. Within topics, terms are ranked by their FREX scores (weighted at 0.5), which balance the frequency with which a term occurs in a topic with the term's exclusivity to the respective topic \citep{airoldi2016improving}. 

\begin{figure}
\centering
\includegraphics[width=1\columnwidth]{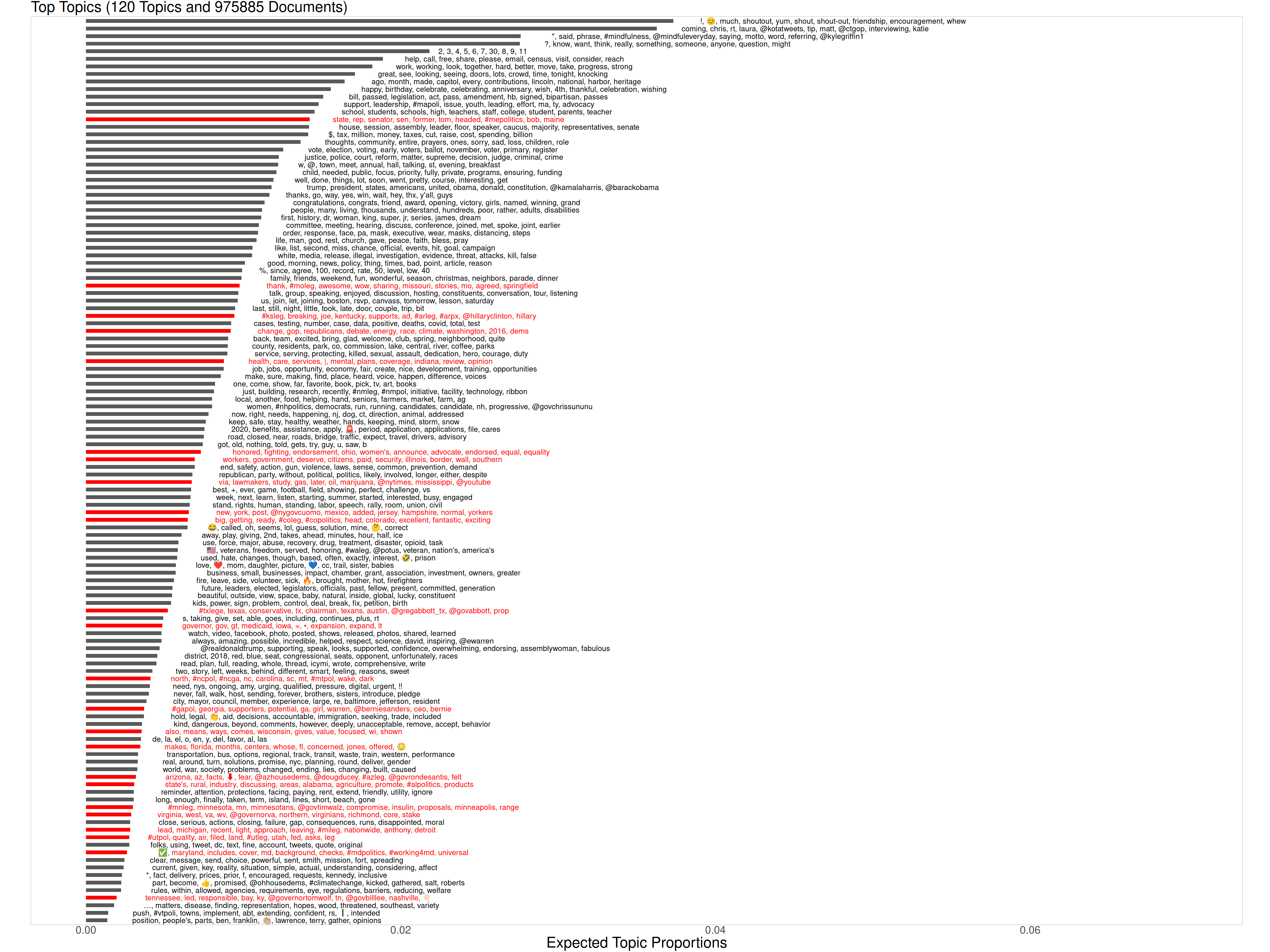}
\caption{120 topics STM model. Each document is the text of a single tweet.}
\label{fig:twt1}
\end{figure}

\begin{figure}
    \centering
    \includegraphics[width=1\columnwidth]{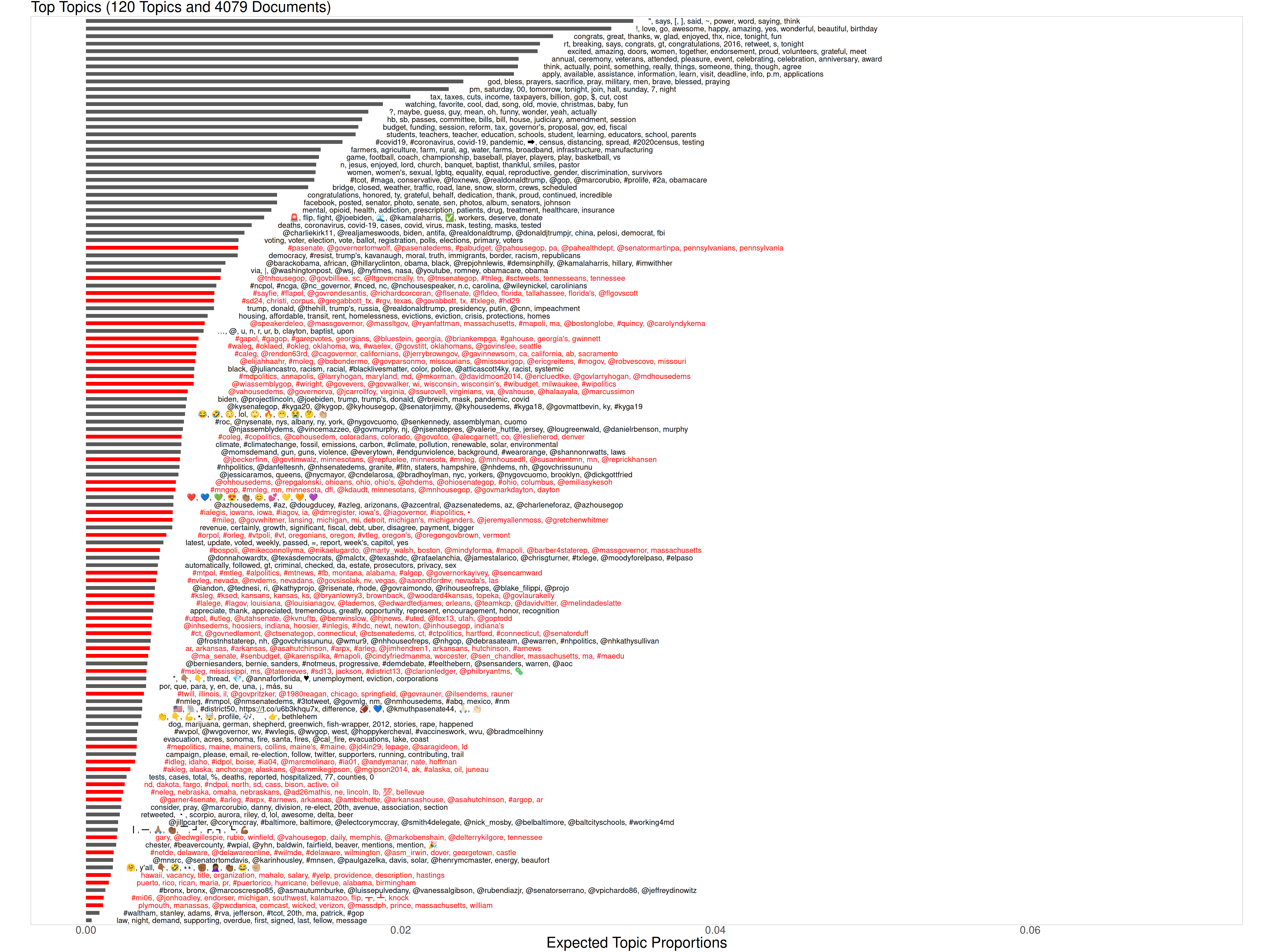}
    \caption{120 topics STM model. Each document is the text of tweets aggregated by legislator.}
    \label{fig:twt2}
\end{figure}

Figures~\ref{fig:twt1} and \ref{fig:twt2} display all of the topics from the 120-topic models fit to the tweet and legislator level document definitions, respectively. The topics are colored in red if at least one of the top ten terms is ``state-related''. 
A topic is considered state-related 
if one of its top ten words is a state name (e.g. `pennsylvania', `massachusetts') or two of its top ten words make up a state name (e.g.``new'' and ``hampshire'' or ``north'' and ``dakota''). We acknowledge that there are state-related topics that do not specifically name the state, so in the online appendix we expand the dictionary of state-related terms to include relevant offline and online vernacular as a robustness check that yields similar results to the main analysis. Furthermore, we posit that legislators have more state-focused discussion due to the nature of their positions. They are responsible for local and state laws and compared to federal level officials state legislators have a more immediate proximity to their constituents \citep{tan2009local, maestas2003incentive, gamm2010broad}. Therefore, legislators are incentivized to prioritize their constituency preferences, which largely includes local and state related policy-making \citep{maestas2003incentive, gamm2010broad}.

The topics we find are highly interpretable; meaning the words in each topic share a coherent theme. The top topics cover, e.g., general sentiments (e.g., happiness/excitement, religion), announcements/deadlines, and information dissemination.  The resulting topics and topic content are relatively consistent for both the 120 and 60-topic models---with some more subtle differences---across the number of topics, so we only present the results of models with 120 topics. Modeling results with 60 topics are presented in the online appendix.

The primary result on which we focus regards topics that are state-related. 
There is a significant difference in state-related topic output when comparing the two document definitions. The topics in tweet-level documents are much less state oriented than those fit to legislator-level documents. This difference shows that document definitions matter in terms of the interpretation of topics. By aggregating to the legislator level, the topics are inferred to organize the documents (legislators) into coherent groupings, which largely correspond to states. With legislator-level documents, we conclude that states figure much more prominently into the topics to which legislators allocate attention in their tweets.

In Table~\ref{tab:tweets} we present the number of topics that include a reference to a state in their top ten tokens in terms of FREX scores. 
The legislator defined topic models feature a greater number of topics that discuss individual states compared to the models with the tweet-based documents.  In each model, the legislator-level document version includes nearly twice as many state-related topics.

\begin{table}[hbt!]
\centering
\caption{Number of topics whose top 10 FREX include state-related words (Sum of proportions of state-related topics): Tweets Document Definition vs. Legislator Document Definition.}
\label{tab:tweets}
\begin{tabular}{lll}
\toprule
&Tweets  &Legislator\\
\midrule
\texttt{60-topics} & 17 (0.17) & 34 (0.31) \\
\texttt{120-topics} & 24 (0.13) & 43 (0.21) \\
\bottomrule
\end{tabular}
\end{table}

In retrospect, this result is not surprising. Since we aggregate documents according to a factor, the legislator who posted the tweet, that is itself geographically partitioned, geographic locations emerge as important discriminating features. We do note, however, that past research that has employed the ``user scheme'' for topic modeling of tweets has justified doing so to infer more interpretable topics \citep[e.g.,][]{li2016suggest,kateb2015classifying}. Systematic substantive differences, such as the one we have identified, have not, to our knowledge, been documented in previous work. 

Our primary interest in the difference between tweet and legislator definitions of the documents is methodological---illustrating that aggregation can change the substantive composition of the topics. However, our findings in this application are substantively significant in the context of an ongoing debate in the literature in state politics. Researchers have studied the degree to which state legislatures are focused on state-related issues versus generic partisan topics \citep[e.g.,][]{garlick2017national,hunt2023local}. Our results indicate that conclusions regarding the prevalence of state-related topics in state legislators' online communications depends on the level at which  documents are defined. The legislator definitions result in twice as many state-related topics as with the tweet definitions. These two numbers can be interpreted as giving lower and upper bound estimates of the relative attention to state-related topics in state lawmakers' tweets. 

To further test if the legislator-level model captures state-specific features of the topic distribution better than the tweet-level model, we assess predictive validity \citep{quinn2010analyze}. Specifically, we test whether regression models using the topic distributions in a legislator's tweets can predict their state. If the legislator-level model better reflects state-specific topics, we expect it to predict a legislator's state more accurately than the tweet-level model. We fit a multinomial logit model where the dependent variable indicates a legislator's state, using 120 covariates—one for each topic proportion in a legislator's tweets. For the legislator-level model, a topic's proportion is given by the estimated proportion in the single legislator-level document; for the tweet-level model, it’s the average proportion across all tweets by the legislator. In both cases, each legislator's covariates sum to 1. We fit two multinomial logit models using each topic distribution as covariates, with 4,079 observations in each dataset.

We compare the predictive performance of these covariates in two ways. First, we fit both models to the full dataset and compare in-sample fit using AIC. Second, we randomly split the data into training and test sets, with 1000 observations or about 25 \% allocated to training and 3079 observations or about 75 \% to testing, respectively, and compare the two models in terms of their accuracy in predicting the legislator's state. We fit the model to the training observations and predict the state in the test observations. The proportion of correct predictions in the test data indicates accuracy.  
Table~\ref{tab:preds} presents the AIC scores for each multinomial logit model. The legislator-level model has a substantially lower AIC score, indicating that the legislator-level model more accurately identifies state-specific discussion than with the tweets-level model. In the out-of-sample test, the tweet-level model accurately predicts a legislator’s state correctly 59\% of the time, compared to 86\% for the legislator-level model. These results show that the legislator document definition better captures state-related discussion than the tweet-level definition.

\begin{table}[hbt!]
\centering
\caption{AIC scores and out-of-sample predictive accuracy (proportion of accurate state predictions for each legislator) for 120-Topic model: Tweets Document Definition vs. Legislator Document Definition.}
\label{tab:preds}
\begin{tabular}{lll}
\toprule
&Tweets  &Legislator\\
\midrule
\texttt{AIC Score} & 23,489.63 & 12,032.07 \\
\texttt{Accuracy} & 59.2\%\ & 85.7\%\ \\
\bottomrule
\end{tabular}
\end{table}

In the online appendix we include a robustness check in which we evaluate whether our findings are solely attributable to the fact that aggregation makes documents longer. 
We find that randomly aggregating tweets does not result in as high a number of state-related topics as when we aggregate to the legislator level. Aggregation by legislator results in a greater number of state-related topics and greater proportion of state-related topics across both 60-topic and 120-topic models.

\section{Discussion and Recommendations}

We find that aggregating documents at the legislator level results in topic distributions that are more indicative of legislators's states. Why is this? We see the concept of entropy as 
central to the behavior of the topic models that we have identified. Interpretable topic models are characterized by low-entropy distributions of words across topics \citep{koltcov2019estimating,cade2010using}. Indeed, topics are just clusters of words, and if the entropy in word-topic/cluster distributions is high enough, these clusters are not interpretable as ``topics''.  The maximum entropy distribution of any word across topics would be a uniform distribution in which a word was equally likely to occur in each topic. Low entropy word distributions are those in which individual words occur with high likelihood in a small number of topics. Since references to individual states are relatively rare in any given tweet, but are effective at differentiating individual legislators from each other, when we aggregate the tweets up to the legislator level, we change the role of state references in constructing low entropy topic distributions. When aggregated up, features of text that are ineffective at differentiating individual short-text documents, but effective at differentiating the units to which aggregation has been done, will play a greater role in defining the topic distributions under aggregated documents.

Topic models are used on a variety of text data sources in the study of legislative politics including, e.g.,  twitter \citep{payson2022using, butler2023male}, lobbying reports \citep{kim2021state}, and legislative bills \citep{callaghan2021bill, kroeger2022groups, dehart2023reserved}. It is common for researchers to optimize the preprocessing steps and parameter settings for a given application \citep[e.g.,][]{gilardi2021policy,roberts2014structural}. With topic modeling this can include selecting the number of topics to infer, or experimenting with different prior distribution specifications. We introduce the level of document aggregation as another dimension for researchers to consider when working with short text. By aggregating documents up to some relevant unit of analysis, researchers can focus their inferences on features in the data that differentiate the units from each other. We see the level of aggregation as another setting in the text preprocessing pipeline. 

The purpose of our research is to highlight that topic models perform differently with longer length documents but not necessarily better. In some cases, there will be an optimal level of aggregation; meaning document aggregation can reveal more meaningful results based on the specific research question and features of the documents themselves. When researchers use topic models to measure, e.g., specific features of the topics in accounts' tweets \citep{munger2019elites}, the level of aggregation can be manipulated to achieve the best measurement performance. In other cases, e.g., when topic models are used to explore the organization of content in a new corpus \citep{montiel2021language}, the level of aggregation should simply be used for further exploration. Dis-aggregating documents to, e.g., sentence \citep{daubler2012natural} or paragraph \citep{gilardi2021policy} level, is another approach to varying the degree of document aggregation in text analysis---an approach that researchers have found to be valuable in optimizing the results of content analysis. In the case of text embedding models, which can be used as a basis for topic modeling, the ``context window'' parameter serves to (dis-)aggregate text \citep{arseniev2022integrating}. Also if there are multiple authors for a single document then we recommend replicating the documents across each author.

In nearly all topic modeling projects it would make sense for researchers to consider the level of document aggregation right along with the other hyperparameters (e.g., prior parameter values, number of topics) 
We recommend that researchers approach (dis-)aggregation of documents as a two-step process. First, if the research objective is primarily descriptive or exploratory, such that the aim of the research is to accurately summarize the topics in the corpus, we recommend that researchers identify 2--3 levels of aggregation, and distinguish between topic areas that are prevalent regardless of the level of aggregation and those that emerge at just one level of aggregation. These levels of aggregation should fit the constraints of the analytical framework in the specific application (e.g., if researchers are studying change in rhetoric over time, it is not feasible to aggregate over time). Second, if there is a specific confirmatory and/or predictive objective of the research (e.g., predicting election polls \citep{beauchamp2017predicting}), we recommend that researchers define 2--3 levels of aggregation, and select the level that performs best in the downstream task. This could be done using a cross-validation experiment in which topic labels are applied to texts that are not used to infer the topics \citep{hu2020identification}. It is tempting to consider using common measures of topic model fit---something like held-out likelihood \citep{wallach2009evaluation}---to directly compare the performance of models at different levels of aggregation. Unfortunately, the aggregation step changes the outcome variable by combining word occurrences across documents, making likelihood calculations incomparable across levels of aggregation.  An opportunity for future research would be to develop measures of topic model fit that are comparable across levels of aggregation.


\section*{Funding Statement}

This research was supported by grants from the National Science Foundation (2318460).

\bibliographystyle{plainnat}
\bibliography{references}

\end{document}


\maketitle

\vspace{-3cm}

\section{BERTopic Results}

Since large language model (LLM) approaches are rapidly emerging in the text-as-data toolkit, we want to evaluate the robustness of our findings against a transformer-based approach. BERTopic uses a BERT-based language model to represent each document as an embedding vector. Then, the embeddings are represented in fewer dimensions by running a dimensionality reduction algorithm. Then, similar documents are clustered together using a clustering algorithm. Lastly, each cluster is represented with a set of words using an adjusted TF-IDF algorithm that treats each cluster as a document. We use BERTopic's default embedding model Sentence-BERT \citep{reimers-2019-sentence-bert} and use its default dimensionality reduction algorithm UMAP \citep{mcinnes2018umap}. For clustering, we use k-means clustering which allows us to explicitly create 60 and 120 topic models, replicating our STM analysis closely. Unlike the STM models, we do not remove any infrequent or stop words during pre-processing as they are important for the transformer architecture to create accurate embeddings. However, in line with BERTopic documentation's recommendation, we remove the stop words from the topic representations so that the top 10 words generated by the adjusted TF-IDF that represent a topic are not stop words. We apply the same definition of a state-related topic as we did in STMs on these topic representations.

One constraint imposed by any transformer-based model is that they have a maximum number of tokens they can process. In the case of BERTopic's default model, this number is 256 tokens.\footnote{BERTopic's default embedding model, as of this writing, is \texttt{all-MiniLM-L6-v2} Sentence-Bert model (https://huggingface.co/sentence-transformers/all-MiniLM-L6-v2). The model automatically truncates any text longer than 256 tokens. Even though it is possible to override this number, we stick to working with the model's default setting as it has been optimized during training for semantic similarity in shorter text.} This makes the model ideal for tweets, but the model cannot directly process many of the documents aggregated at the legislator level. Therefore, we create embeddings at the legislator level by taking the mean of all of that legislator's tweets' embeddings, and reducing them down to a single embedding. Every other step remains the same.

\begin{figure}
\centering
\includegraphics[width=1\columnwidth]{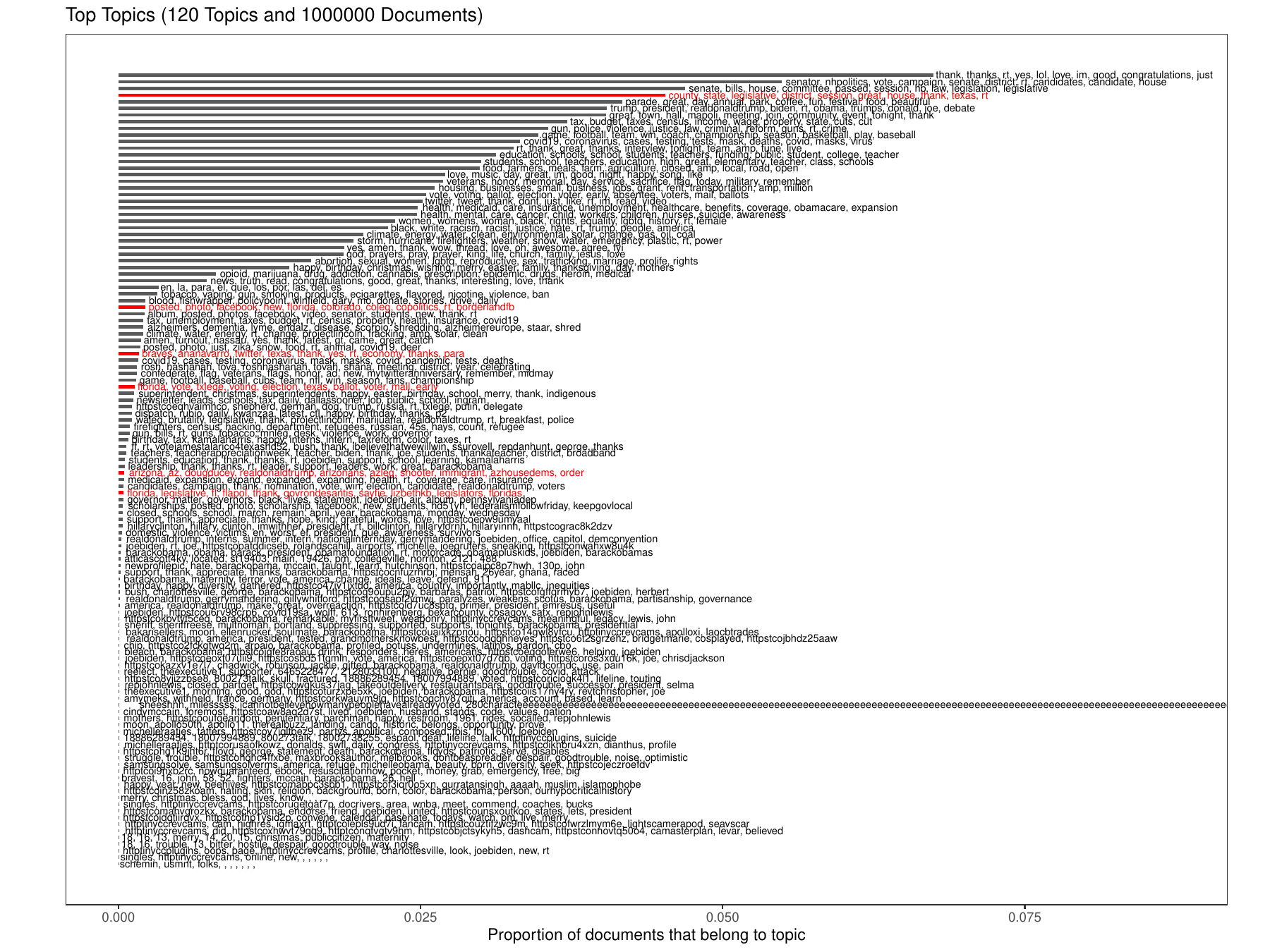}
\caption{120 topics BERTopic model. Each document is the text of a single tweet.}
\label{fig:bertopic_120_tweet_level}
\end{figure}

\begin{figure}
    \centering
    \includegraphics[width=1\columnwidth]{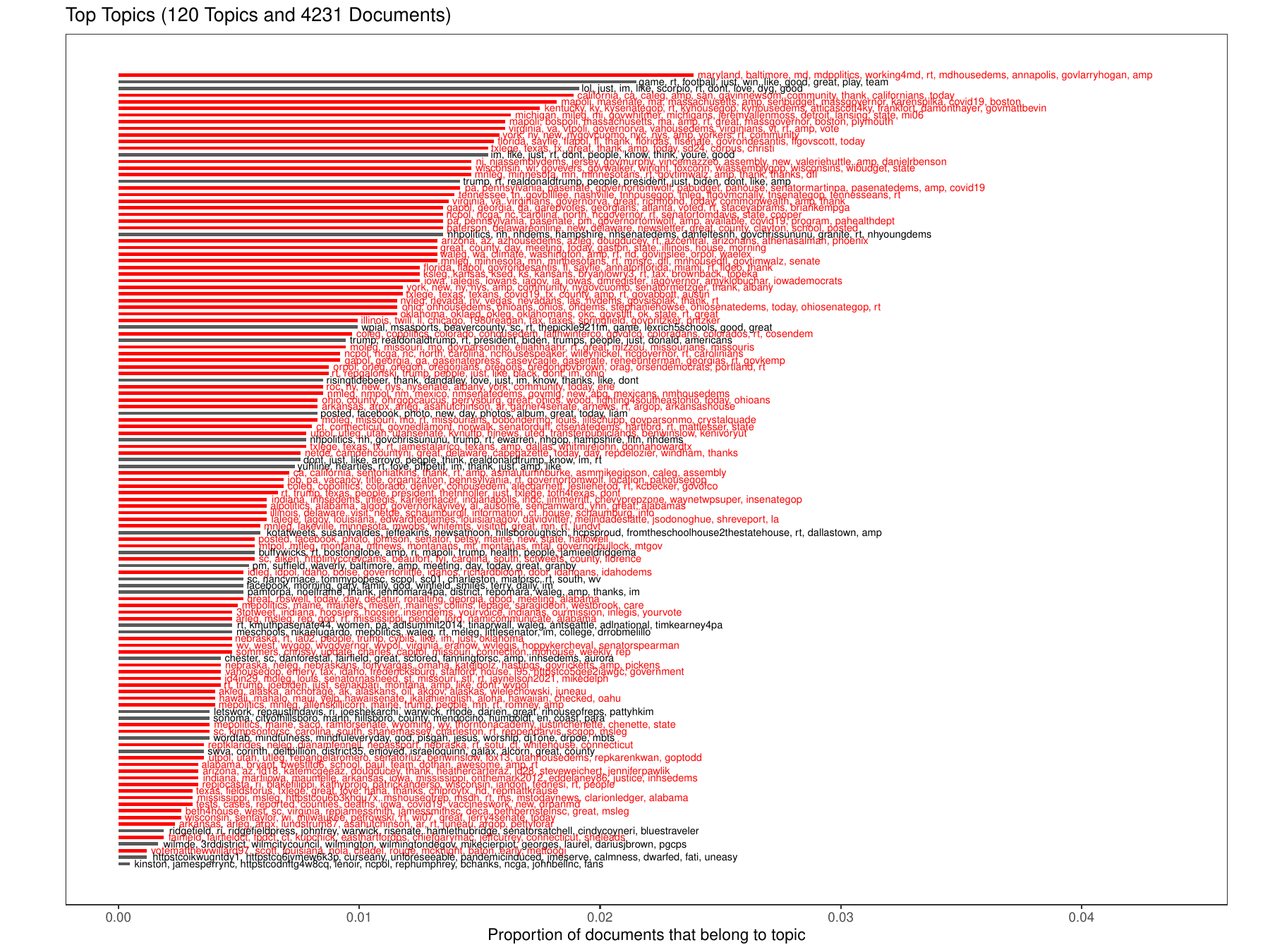}
    \caption{120 topics BERTopic model. Each document is the text of tweets aggregated by legislator.}
    \label{fig:bertopic_120_legislator_level}
\end{figure}

We summarize our results from the four BERTopic models in Table~\ref{tab:bertopic_result_summary}. The results show the emergence of the same pattern in an even starker way. For both 60 and 120 topic models, aggregating the topics at the legislator level results in much more state mentions in topic representations.
Figures~\ref{fig:bertopic_120_tweet_level} and \ref{fig:bertopic_120_legislator_level} visualize the topic representations, highlighting the topics that are state-related.

\begin{table}[hbt!]
\centering
\caption{BERTopic: Number of topics whose topic representations include state-related words (Proportions of documents that belong to state-related topics): Tweets Document Definition vs. Legislator Document Definition.}
\label{tab:bertopic_result_summary}
\begin{tabular}{lll}
\toprule
 &Tweets  &Legislator\\
\midrule
\texttt{60-topics} & 4 (0.01) & 47 (0.73) \\
\texttt{120-topics} & 6 (0.05) & 91 (0.79) \\
\bottomrule
\end{tabular}
\end{table}

Lastly, we also replicate the predictive validity experiment with the BERTopic models. 
We again fit multinomial logistic regression models in which the outcome is the legislator's state and the predictors are probabilities of belonging to one of 120 topics.
For the legislator-level document definition, every legislator is necessarily placed into one topic by the k-means clustering algorithm, which means that, for a given document, one of the topics is 1 and every other covariate is 0.
For the tweet-level document definition, each tweet by a legislator may belong to a different topic. This is converted to a probability distribution over all topics by treating the relative proportion of a legislator's tweets that belong to a topic as the probability of that topic, which means the value for each topic varies between 0 and 1 and all the topic covariates for a single legislator sum up to 1.

To assess model quality, we first fit the models to the entirety of the data and use AIC scores to compare the tweets-level and legislator-level document definitions. Secondly, we randomly split the data into training and test sets ($\sim$\%75 to $\sim$\%25), train the model using the training set, and compare the accuracy scores on the test set. Both sets of results are presented in Table~\ref{tab:preds_bertopic}. The lower AIC score and higher predictive accuracy in the test set when using topics generated by the legislator document definition show that these topics are better at the task of predicting which state a legislator belongs to.

\begin{table}[hbt!]
\centering
\caption{AIC scores and out-of-sample predictive accuracy (proportion of accurate state predictions for each legislator) for 120-Topic model: Tweets Document Definition vs. Legislator Document Definition.}
\label{tab:preds_bertopic}
\begin{tabular}{lll}
\toprule
 &Tweets  &Legislator\\
\midrule
\texttt{AIC Score} & 30,557.13 & 20,939.35 \\
\texttt{Accuracy} & 14.2\%\ & 66.1\%\ \\
\bottomrule
\end{tabular}
\end{table}

\section{Application on the Wikipedia corpus}
To evaluate the robustness of our findings regarding the effects of document aggregation in topic models, we analyze a second corpus---Wikipedia articles about U.S. citizens. Wikidata—Wikipedia’s sibling project that stores structured data—is a useful tool to query persons in Wikimedia projects using a set of biographical criteria and to retrieve data on them \citep{yalcin_2022}. This includes place of birth and links to their Wikipedia page, which is what we use. Using Wikidata, we find human beings who are U.S. citizens born between January 1, 1920, and December 31, 2000, who have exactly one English Wikipedia page associated with them, and whose birthplace information is recorded in Wikidata. This results in links to 191,622 Wikipedia pages and birthplace information associated with each person. We then scrape the text from each Wikipedia page. Each birthplace is recorded as a unique Wikidata entity id, which means we can exactly match and aggregate by this birthplace id—instead of having to extract the birthplace from the text itself. 

This corpus presents a well-structured robustness test in the context of our analysis of legislator tweets. Wikipedia articles are longer and more structured than tweets. However, the process of aggregating by birthplace is comparable to aggregating tweets by legislator in that birthplaces are largely partitioned into the U.S. states, except for those that are in U.S. territories or outside the U.S. If our findings are robust, we expect to see the same pattern, with significantly more state-related topics in the model fit to documents aggregated by birthplace than in the model of article-level documents.

We use two document definitions for the topic model. 
In the first definition, each Wikipedia page is a document.
In the second definition, each Wikipedia page is aggregated by birthplace.
This birthplace aggregation is less likely to bear practical application than aggregating tweets to the user level.
The objective in the current analysis is simply to evaluate whether an intermediate-level of aggregation yields topics that are more closely related to states---replicating the patterns from our analysis of legislators' tweets.
A small number of people have more than one birthplace associated with them (e.g. the city and state are recorded separately in Wikidata).
In that case, that document is duplicated and aggregated separately into all of those birthplace entities mentioned---i.e. the same text becomes a part of both documents.
This gives us 17,953 documents associated with 17,953 places.
In this definition, each document contains the Wikipedia pages of people who were born in that place.
We then fit 120-topic and 60-topic models to both sets of documents, using both the STM and BERTopic frameworks.

\subsection{Wikipedia Corpus STM Models Results}

Table~\ref{tab:wikipedia_automated} shows the number of state-related topics in each STM model, determined using the same dictionary-based approach described in the main text and again using the top 10 words based on the FREX scoring algorithm with a ``frexweight'' of 0.5.
The results show that the number of state-related topics is significantly boosted when the pages are aggregated by birthplace.

\begin{table}[hbt!]
\centering
\caption{Number of topics whose top 10 FREX words are state-related (Sum of proportions of state-related topics): Individual Wikipedia pages vs. Aggregation by birthplace. Aggregation by birthplace results in a greater number of state topics.}
\label{tab:wikipedia_automated}
\begin{tabular}{lll}
\toprule
 &Individual Pages  &Birthplace\\
\midrule
\texttt{60-topics} & 8 (0.1) & 21 (0.14) \\
\texttt{120-topics} & 22 (0.11) & 40 (0.13) \\
\bottomrule
\end{tabular}
\end{table}

\begin{figure}
\centering
\includegraphics[width=1\columnwidth]{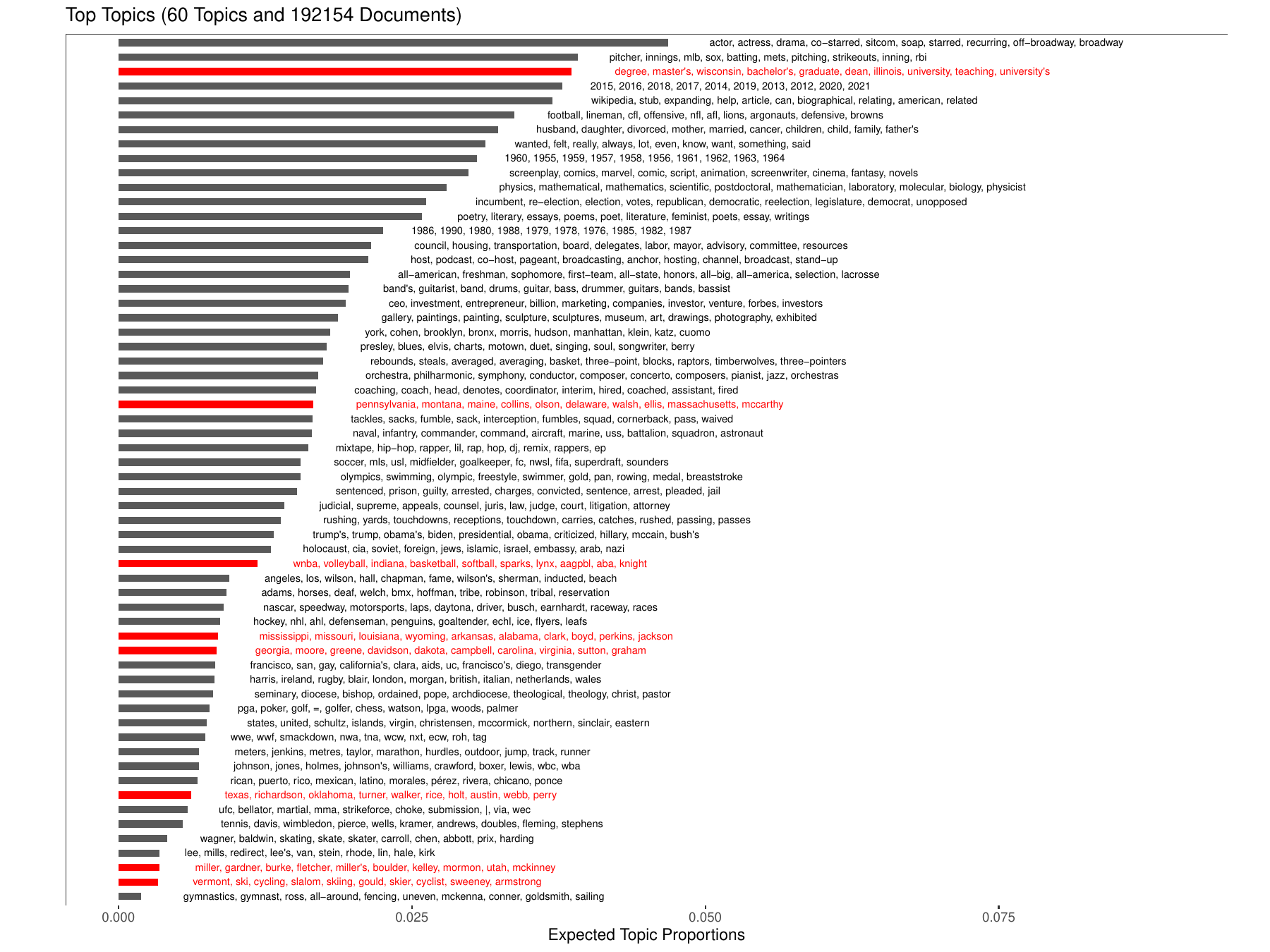}
\caption{60 topics STM model. Each document is text from one Wikipedia page. Model uses Wikipedia corpus.}
\label{fig:wiki1}
\end{figure}

\begin{figure}
\centering
\includegraphics[width=1\columnwidth]{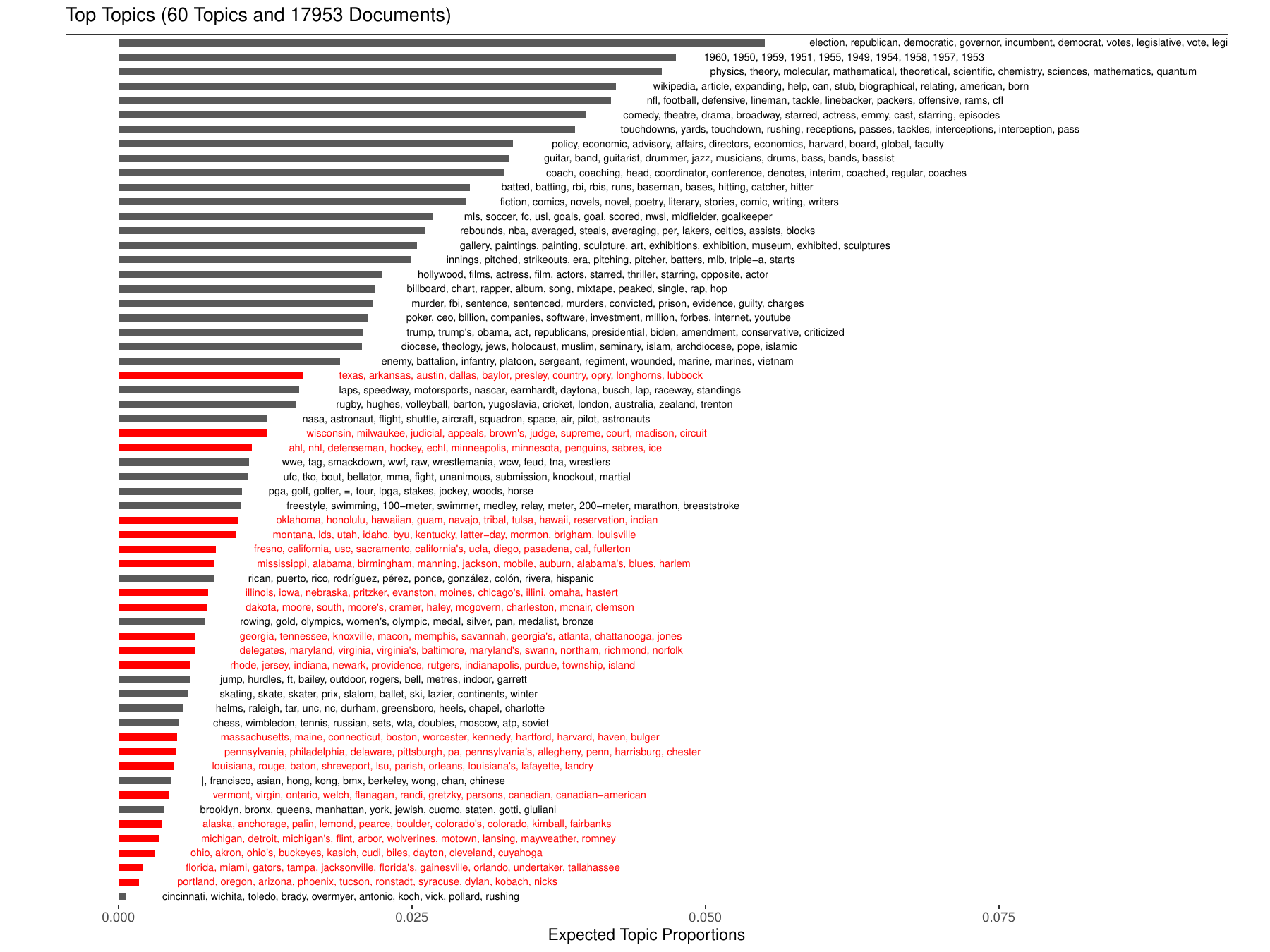}
\caption{60 topics STM model. Each document is associated with one birth place. Model uses Wikipedia corpus.}
\label{fig:wiki2}
\end{figure}

\begin{figure}
\centering
\includegraphics[width=1\columnwidth]{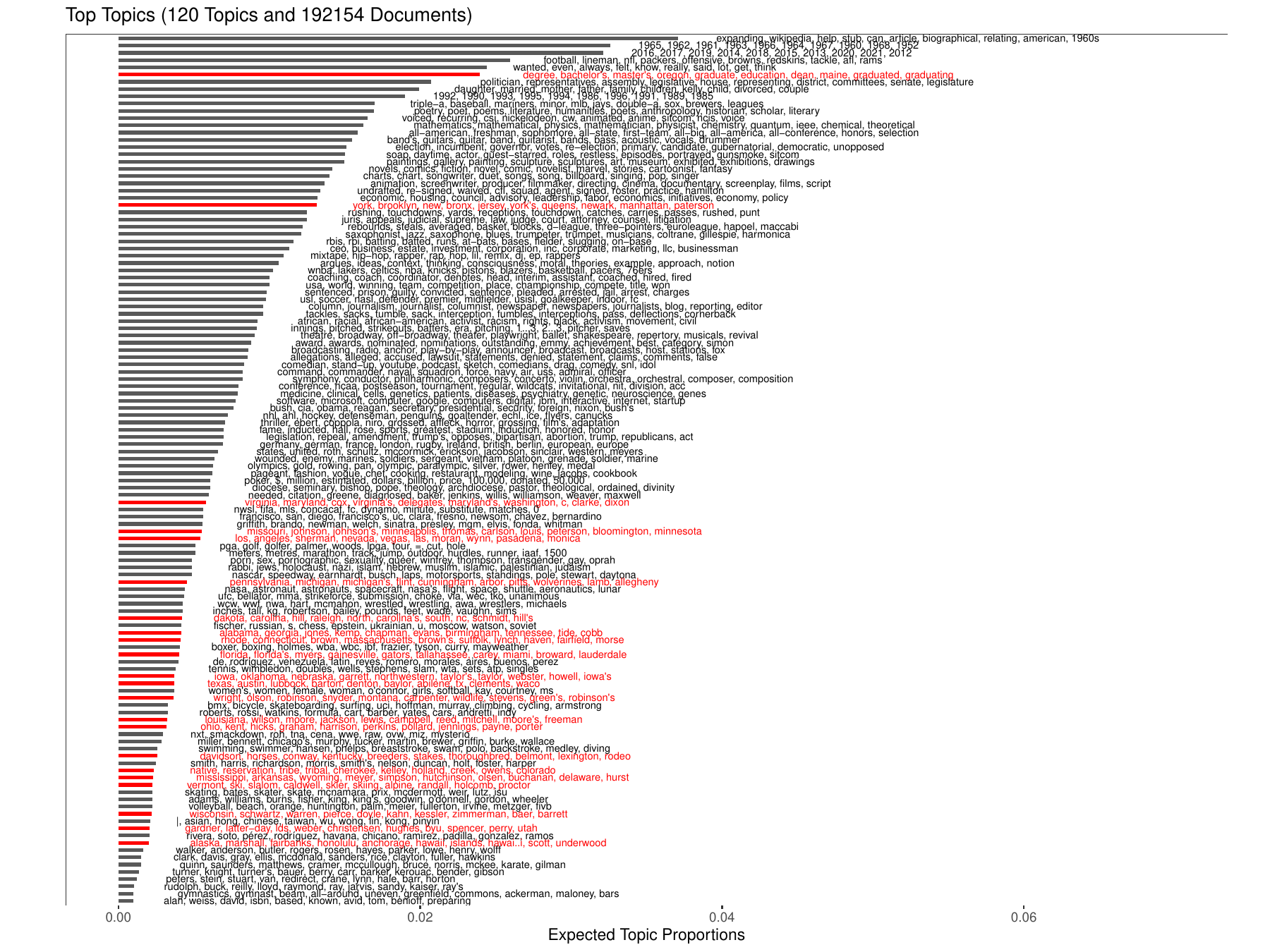}
\caption{120 topics STM model. Each document is text from one Wikipedia page. Model uses Wikipedia corpus.}
\label{fig:wiki3}
\end{figure}

\begin{figure}
\centering
\includegraphics[width=1\columnwidth]{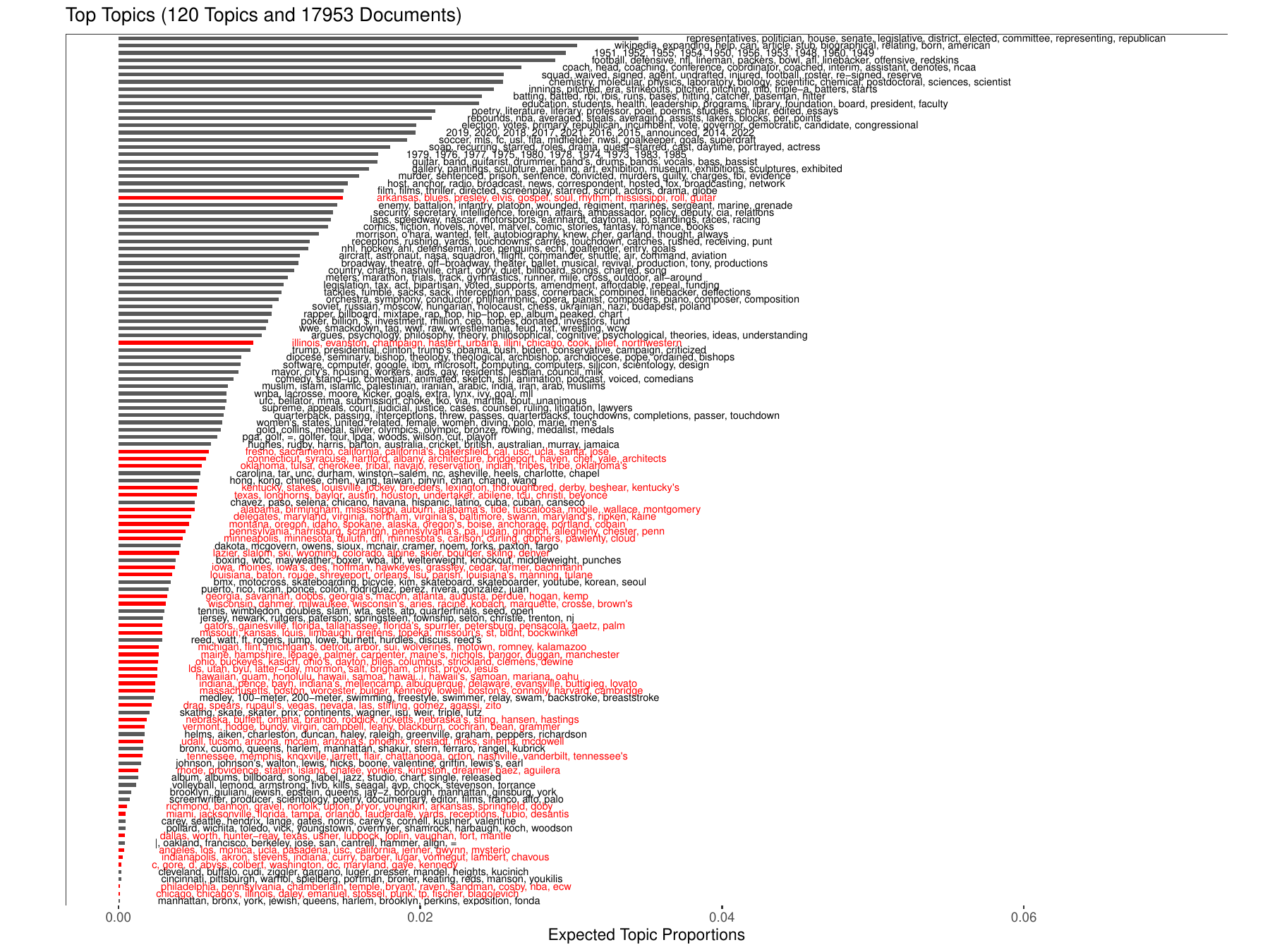}
\caption{120 topics STM model. Each document is associated with one birth place. Model uses Wikipedia corpus.}
\label{fig:wiki4}
\end{figure}

Figures \ref{fig:wiki1}, \ref{fig:wiki2}, \ref{fig:wiki3}, and \ref{fig:wiki4} show the expected topic proportions for each topic in each model from the Wikipedia corpus.
Each topic has its top 10 FREX words next to them.
The topics in red are the topics that are state-related.

\subsection{Wikipedia Corpus BERTopic Models Results}

Table~\ref{tab:wikipedia_bertopic_results} shows the number of state-related topics in each BERTopic model, determined using the same dictionary-based approach described in the main text and again using the top 10 words based on the topic representation that BERTopic generates.
As with STMs, the results show that the number of state-related topics is significantly boosted when the pages are aggregated by birthplace.

\begin{table}[hbt!]
\centering
\caption{Number of topics whose topic representations are state-related (Proportions of documents that belong to state-related topics): Individual Wikipedia pages vs. Aggregation by birthplace. Aggregation by birthplace results in a greater number of state topics.}
\label{tab:wikipedia_bertopic_results}
\begin{tabular}{lll}
\toprule
 &Individual Pages  &Birthplace\\
\midrule
\texttt{60-topics} & 6 (0.07) & 14 (0.30) \\
\texttt{120-topics} & 10 (0.06) & 35 (0.40) \\
\bottomrule
\end{tabular}
\end{table}

\begin{figure}
\centering
\includegraphics[width=1\columnwidth]{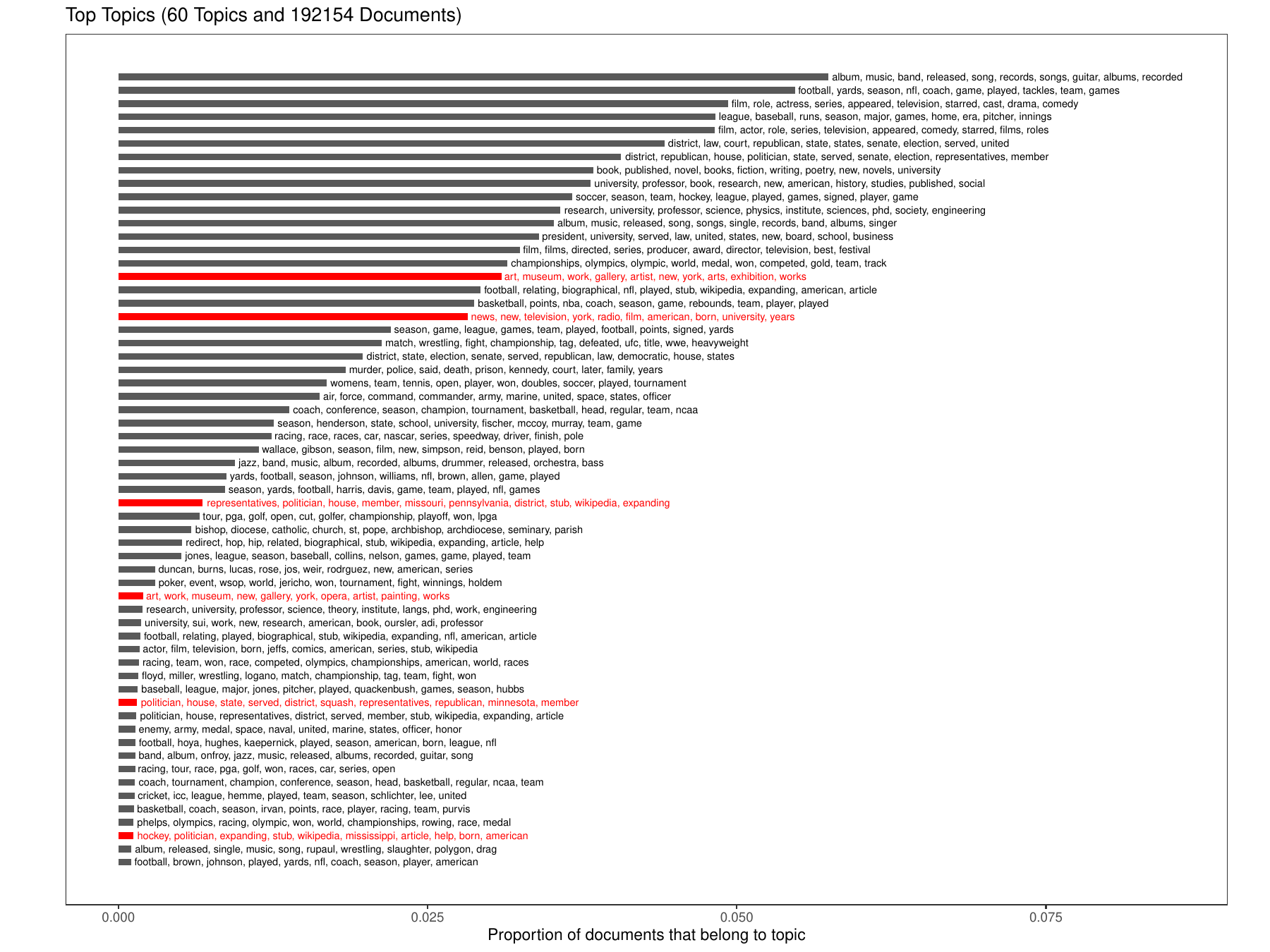}
\caption{60 topics BERTopic model. Each document is text from one Wikipedia page. Model uses Wikipedia corpus.}
\label{fig:wiki_bertopic_1}
\end{figure}

\begin{figure}
\centering
\includegraphics[width=1\columnwidth]{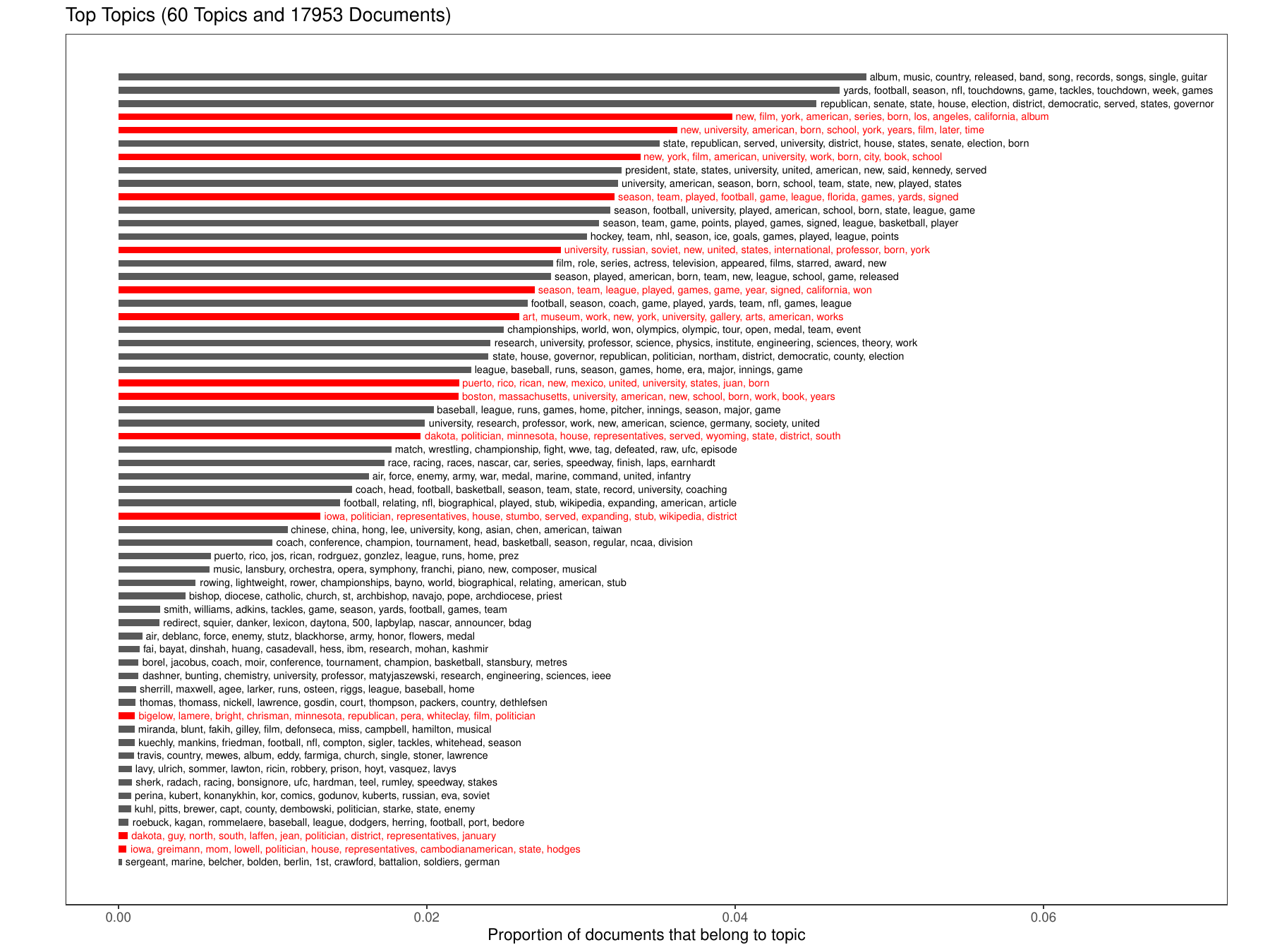}
\caption{60 topics BERTopic model. Each document is associated with one birth place. Model uses Wikipedia corpus.}
\label{fig:wiki_bertopic_2}
\end{figure}

\begin{figure}
\centering
\includegraphics[width=1\columnwidth]{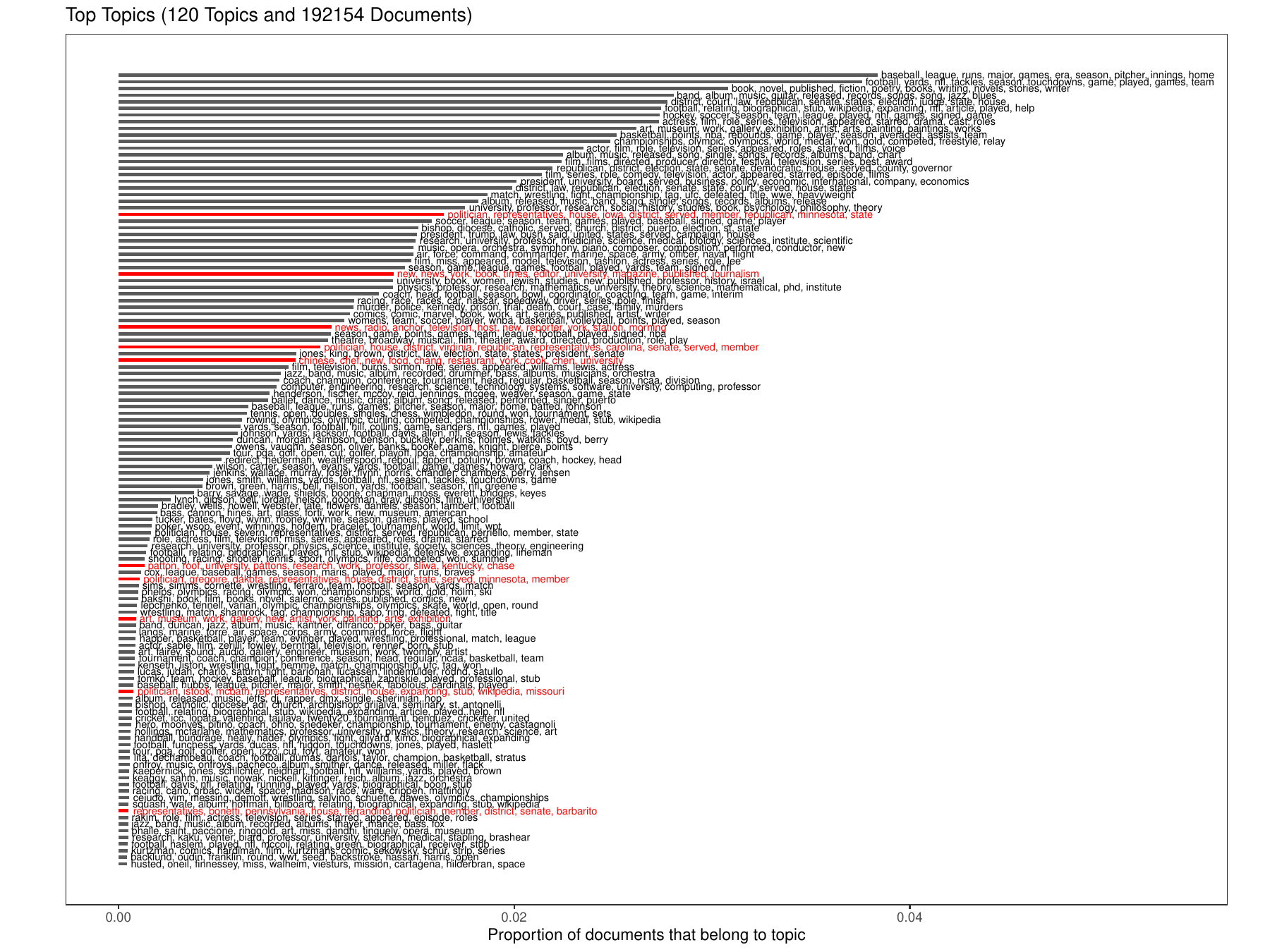}
\caption{120 topics BERTopic model. Each document is text from one Wikipedia page. Model uses Wikipedia corpus.}
\label{fig:wiki_bertopic_3}
\end{figure}

\begin{figure}
\centering
\includegraphics[width=1\columnwidth]{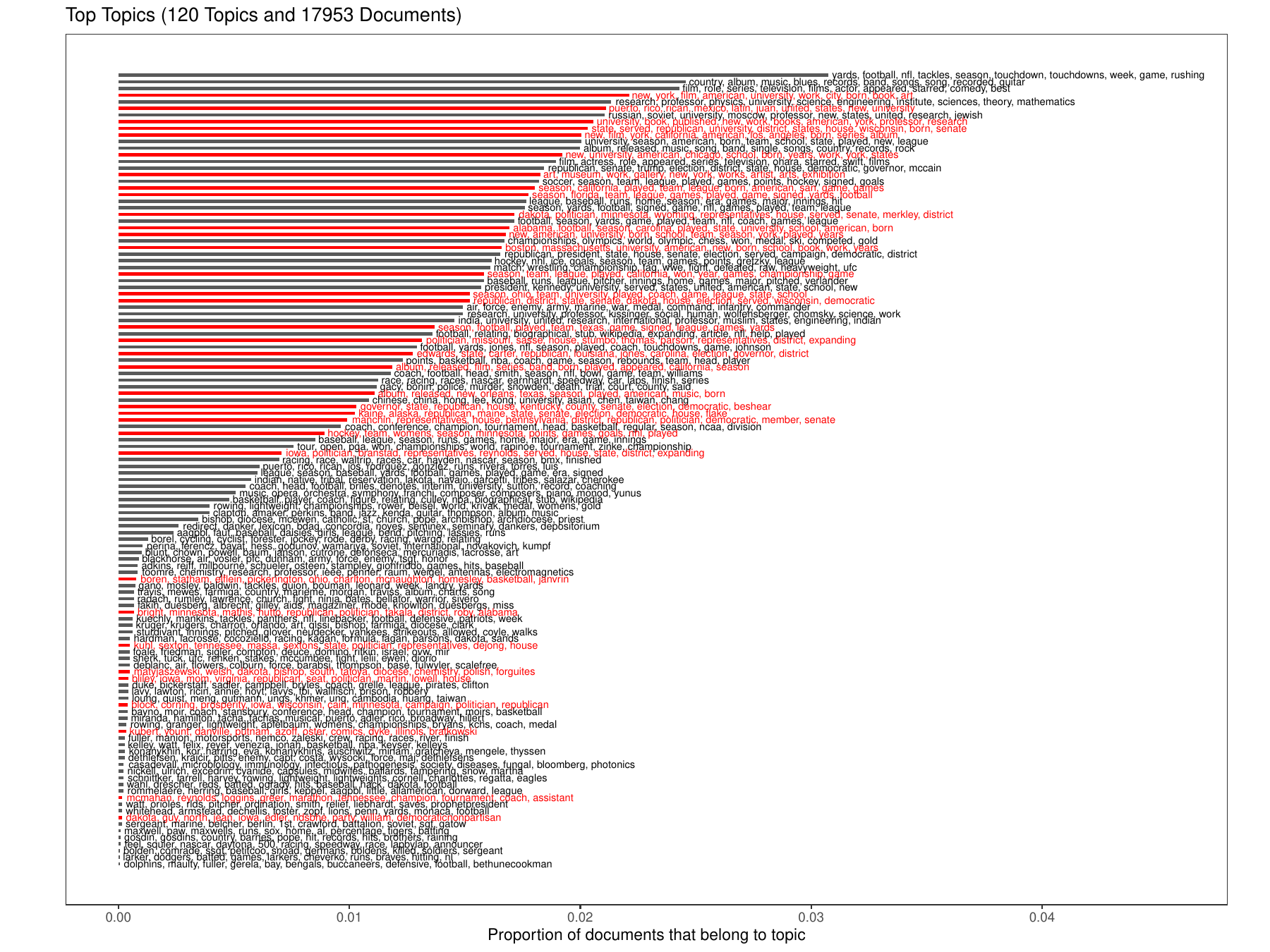}
\caption{120 topics BERTopic model. Each document is associated with one birth place. Model uses Wikipedia corpus.}
\label{fig:wiki_bertopic_4}
\end{figure}

Figures \ref{fig:wiki_bertopic_1}, \ref{fig:wiki_bertopic_2}, \ref{fig:wiki_bertopic_3}, and \ref{fig:wiki_bertopic_4} show the expected the number of documents that belong to each topic for each BERTopic model from the Wikipedia corpus.
Each topic has its top topic representations next to them.
The topics in red are the topics that are state-related.

\section{Visualizations for the 60-topic models for the Twitter Corpus}

\begin{figure}
\centering
\includegraphics[width=1\columnwidth]{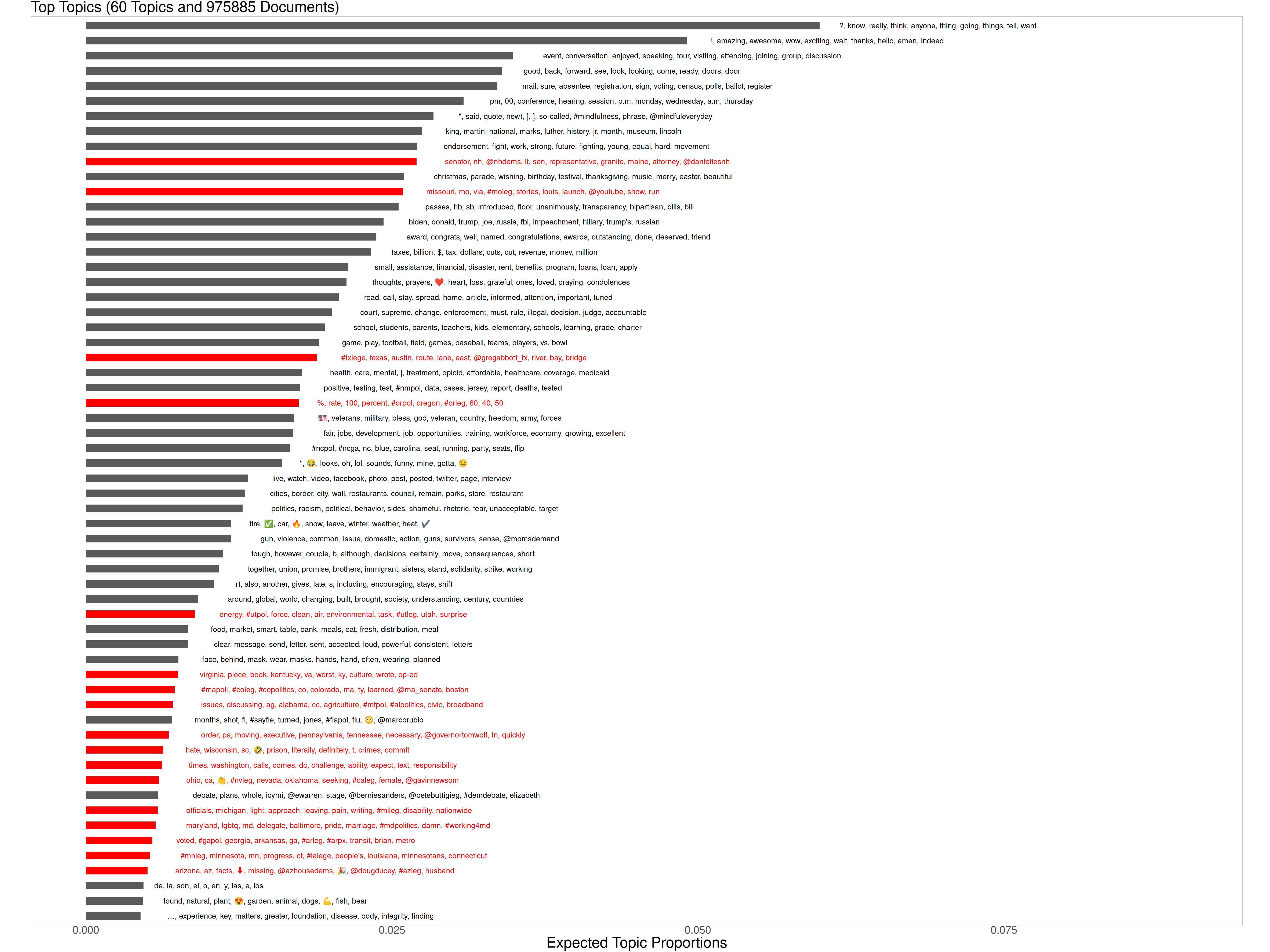}
\caption{60 topics STM model. Each document is the text of a single tweet. Model uses Legislator Twitter corpus.}
\label{fig:smtwt1}
\end{figure}

\begin{figure}
\centering
\includegraphics[width=1\columnwidth]{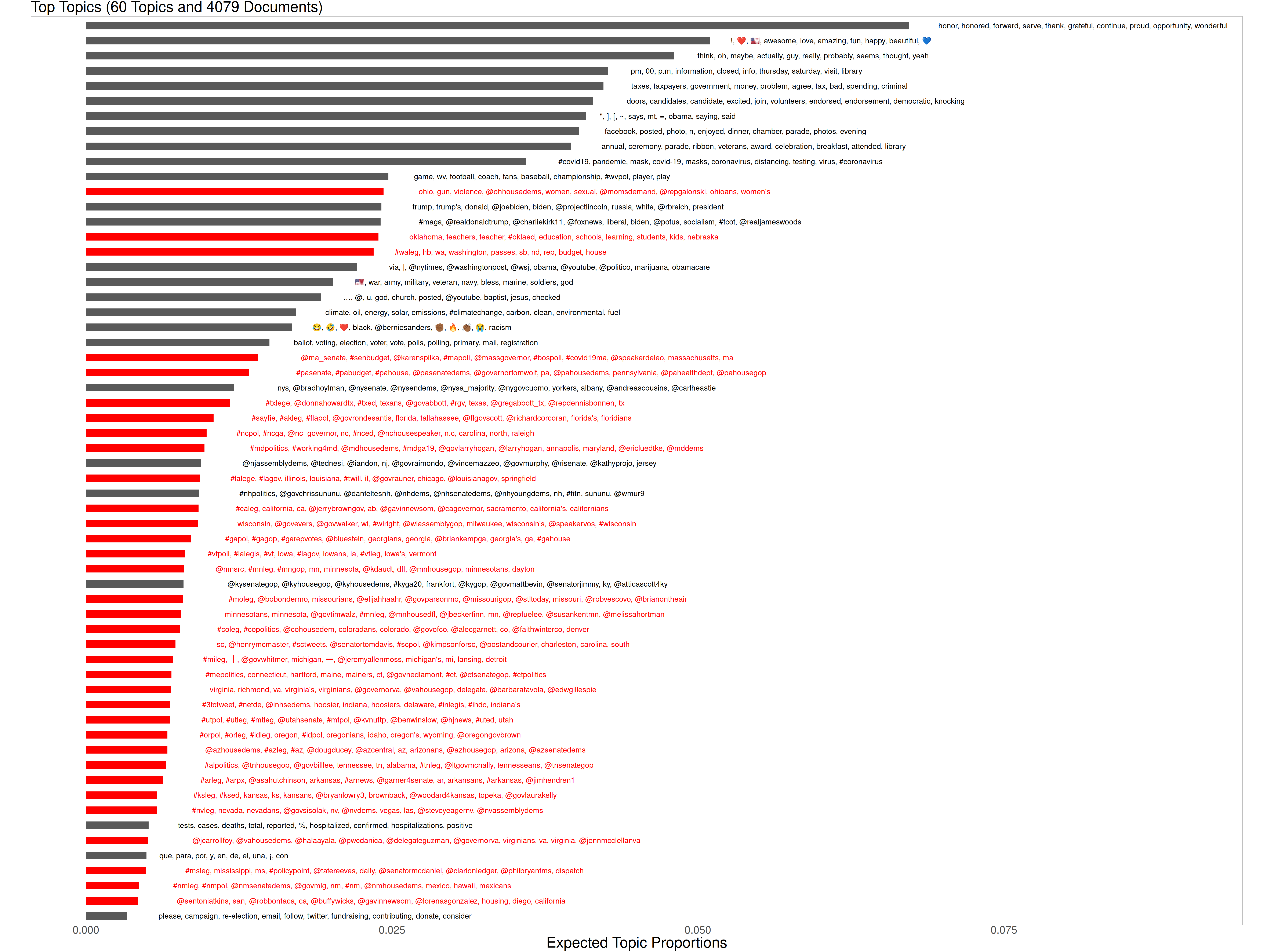}
\caption{60 topics STM model. Each document is the text of tweets aggregated by legislator. Model uses Legislator Twitter corpus.}
\label{fig:smtwt2}
\end{figure}

In the main text, as part of our main analysis, we presented the visualizations for the 120-topic models for the tweet vs. legislator level document definitions.
We present the visualizations for the 60-topic models here.
Figures~\ref{fig:smtwt1} and ~\ref{fig:smtwt2} display all of the topics from the 60-topic models fit to the tweet and legislator level document definitions, respectively.
Similar to the prior figures, each topic is followed by the top 10 FREX words and topics colored in red include words that are state-related.

\section{Dictionary of State-Related Terms}

The analysis in the main text counts state names to determine if a topic is state related. However, on Twitter, legislators will tag their legislatures, states, state party organizations, and colleagues in their tweets, especially when advocating for specific pieces of legislation. As an additional 
measure of whether a topic is state-related, we define a dictionary of tokens that we consider to be state references. We consider a token to be a state reference if (a) it is the name of a state (including if two tokens make up a state name as explained in the main text), (b) it is a state abbreviation, or (c) more than half of that token's occurrence in the Twitter corpus is in documents associated with only one state. This dictionary captures state-related topics that are specific to the vernacular used on Twitter as well as regular offline vernacular. There are 1,830 Twitter-specific tokens that are included in the state-related terms. They include handles of state governors, handles of state legislators, state politics hashtags (e.g., \#ctpolitics, \#mnleg) and names of cities that are highly prevalent within one state (e.g., Harrisburg, Baltimore). In Table~\ref{tab:tweets2} we list the number of state-related topics after including this dictionary of state-related Twitter terms as well as state abbreviations (e.g. AZ, MA etc.). We find that the results are similar to our main findings in that aggregating by legislator yields greater state-related topics even after extending our dictionary of state-related terms.

\begin{table}[hbt!]
\centering
\caption{Number of topics whose top 10 FREX include state-related words: Tweets Document Definition vs. Legislator Document Definition.}
\label{tab:tweets2}
\begin{tabular}{lll}
\toprule
&Tweets  &Legislator\\
\midrule
\texttt{60-topics} & 23 & 42 \\
\texttt{120-topics} & 41 & 71 \\
\bottomrule
\end{tabular}
\end{table}

\section{Incorporating Metadata into STM Models}

Table~\ref{tab:tweets_R1_6} shows the state-related topic output for the legislator-level and tweet-level 60-topic models after including states as a co-variate
for topic prevalence.
The number of state-related topics for both the tweets and legislator document definition models decreases by 2 and 1 respectively. Ultimately, including metadata did not change the results significantly. Including the metadata did, however, significantly increase the computational complexity of estimation, leading to a ten-fold increase in compute time for the tweet-level model, from approximately ten hours without metadata to four days with metadata.

\begin{table}[hbt!]
\centering
\caption{Number of topics whose top 10 FREX words include state-related words: Legislator Document Definition vs. Tweets Document Definition.}
\label{tab:tweets_R1_6}
\begin{tabular}{lll}
\toprule
 &Tweets  &Legislator\\
\midrule
\texttt{60-topics} & 15  & 33  \\
\bottomrule
\end{tabular}
\end{table}

Table~\ref{tab:wikipedia_R1_6} shows the state-related topic counts for the 60-topic Wikipedia corpus models using the state id as a co-variate for topic prevalence.
While the overall pattern is preserved, we see the difference between the document definitions getting somewhat more accentuated.
The model that uses individual Wikipedia articles still has the same number of state-related topics as the model that did not use any metadata about states.
The model that uses the birthplace-aggregated document definition has two more state-related topics than its counterpart that uses no metadata.

\begin{table}[hbt!]
\centering
\caption{Number of topics whose top 10 FREX words include state-related words: Individual Article Document Definition vs. Birthplace Aggregated Document Definition.}
\label{tab:wikipedia_R1_6}
\begin{tabular}{lll}
\toprule
& Individual Pages & Birthplace \\
\midrule
\texttt{60-topics} & 8  & 23  \\
\bottomrule
\end{tabular}
\end{table}

\section{Model Evaluation: Choosing the number of topics}

\begin{figure}[hbt!]
\centering
\includegraphics[width=1\columnwidth]{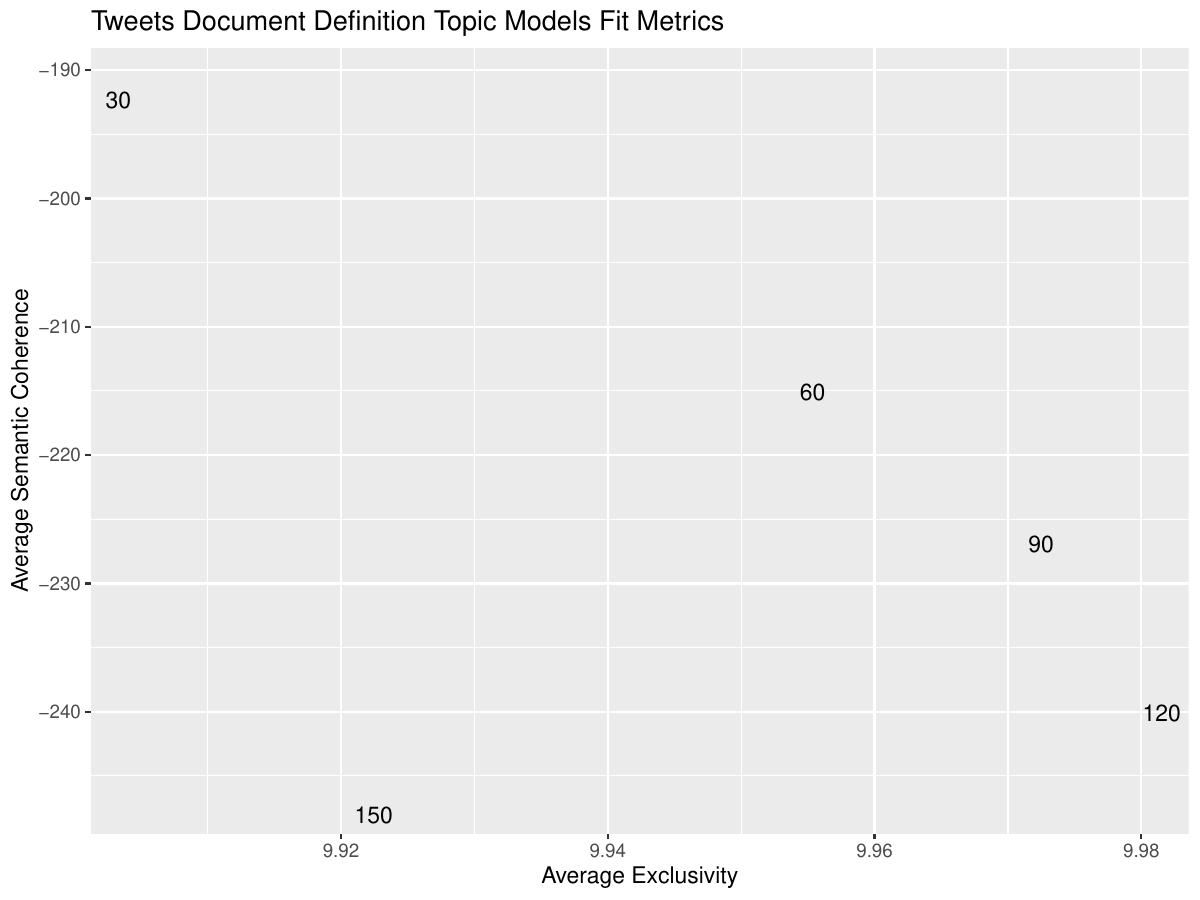}
\caption{The Average Exclusivity vs Average Semantic Coherence values for the topic models of different number of topics, fit with the Tweets document definition}
\label{fig:tweets_coherence_vs_exclusivity}
\end{figure}

\begin{figure}[hbt!]
\centering
\includegraphics[width=1\columnwidth]{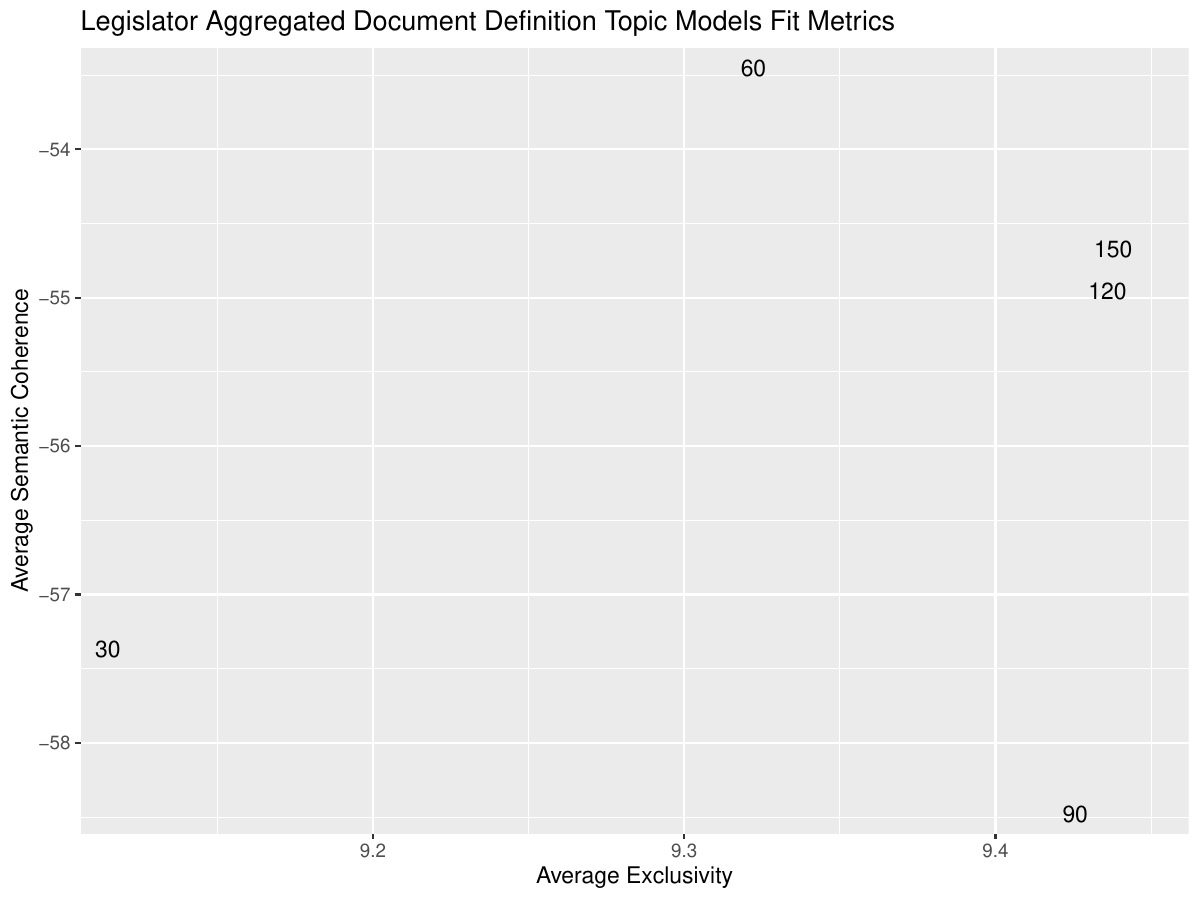}
\caption{The Average Exclusivity vs Average Semantic Coherence values for the topic models of different number of topics, fit with the Legislator document definition}
\label{fig:legis_coherence_vs_exclusivity}
\end{figure}

\begin{figure}[!htb]
    \centering
    \includegraphics[width=\linewidth]{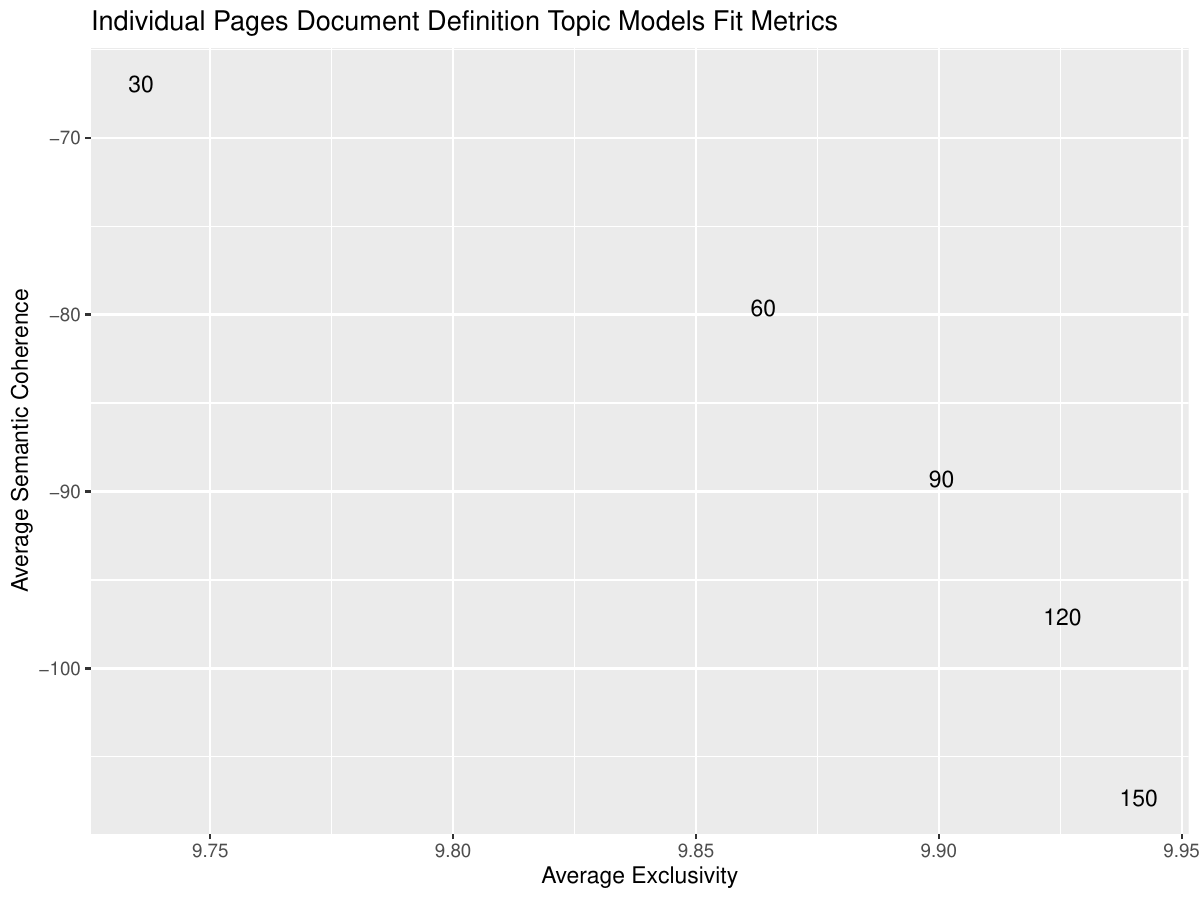}
    \caption{The Average Exclusivity vs Average Semantic Coherence values for the topic models of different number of topics, fit with the individual Wikipedia pages document definition}
    \label{fig:individual_wiki_pages_coherence_vs_exclusivity}
\end{figure}

\begin{figure}[!htb]
    \centering
    \includegraphics[width=\linewidth]{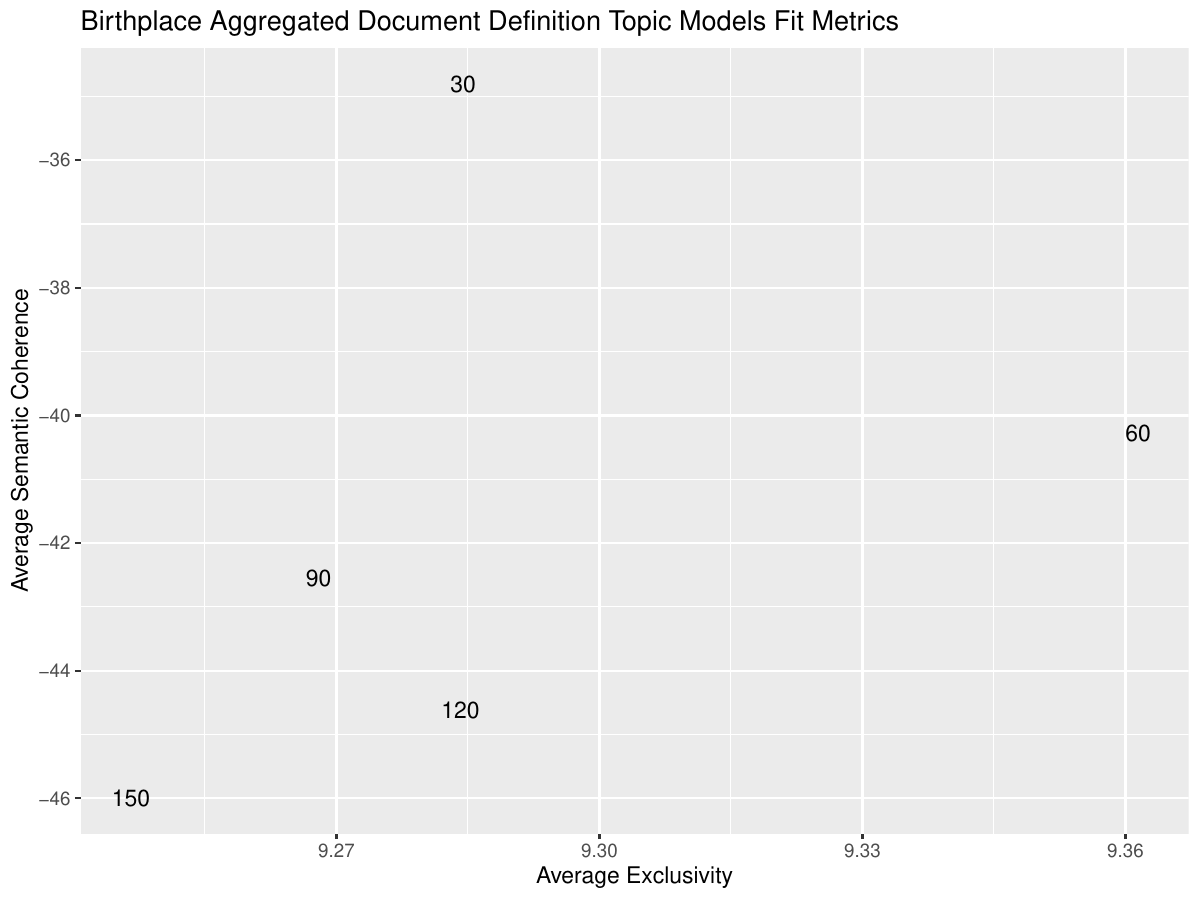}
    \caption{The Average Exclusivity vs Average Semantic Coherence values for the topic models of different number of topics, fit with the birthplace aggregated Wikipedia pages document definition}
    \label{fig:aggregated_wiki_pages_coherence_vs_exclusivity}
\end{figure}

Figures~\ref{fig:tweets_coherence_vs_exclusivity}, ~\ref{fig:legis_coherence_vs_exclusivity}, ~\ref{fig:individual_wiki_pages_coherence_vs_exclusivity} and ~\ref{fig:aggregated_wiki_pages_coherence_vs_exclusivity} show the average exclusivity and average semantic coherence values for 30, 60, 90, 120 and 150 topic models fit to each document definition within each corpus. 

For the models fit to the Twitter corpus, with the Tweets document definition, the 60-topic model provides a good balance of both metrics and the 120-topic model provides the best exclusivity. 
In the legislator document definition, the 60-topic model provides the best semantic coherence and the 120-topic model provides a good average exclusivity.

For the models fit to the Wikipedia corpus, with the individual article document definition, there is a clear monotonic trade-off between exclusivity and semantic coherence.
No number of topics clearly dominates the others on both metrics.
For the birthplace aggregated document definition, the 60-topic model has the most average exclusivity and the 30-topic model has the highest average semantic coherence.

Across all four of our cases---two document definitions across two corpora---there are no models that strictly dominate both the 60 and 120-topic models in terms of both semantic coherence and exclusivity.
Our choice of 60 and 120 topic models is consistent with the recommendation in \citet{roberts2014structural} about selecting a model on the semantic coherence-exclusivity ``frontier.''

\section{Aggregating Documents with Permuted Legislator IDs and Birthplace}

In this section, we do a permutation test to differentiate the impact of aggregating documents by meaningful metadata (legislator, birthplace) and random aggregation.
Aggregating the text affects features of the document definitions in ways beyond the average length of the text. For example, since some legislators tweet more than others, there is more skewness in the number of tokens in each document in the aggregated text (1.66) than in the tweet-level documents (0.74). To conduct this robustness check, we randomly group tweets and Wikipedia pages to construct documents with comparable lengths to those that result from aggregating tweets to the legislator  or birthplace level. 

\begin{figure}[!htb]
    \centering
    \includegraphics[width=0.75\linewidth]{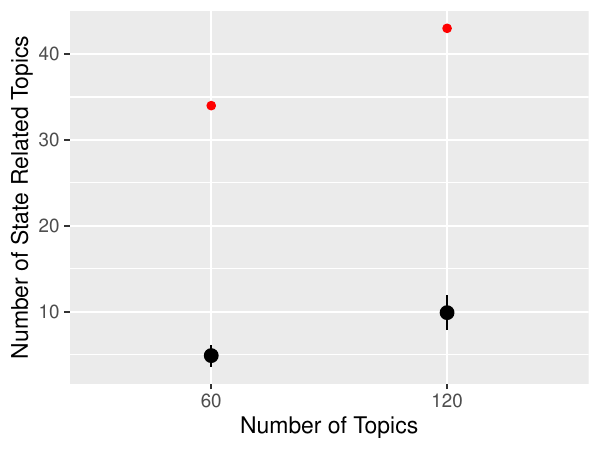}
    \caption{Distribution of state topics of the permuted legislator level models compared to original legislator level models, Single red dot indicates the state-related topic counts of original models, black dot and whisker points indicate the mean state-related topic counts and the 95\% confidence interval around the permuted models.}
    \label{fig:permutations_Twitter}
\end{figure}

To clarify, in practice, we do not recommend randomly aggregating documents and instead choose aggregation that is relevant to the documents structure/research questions. We use this additional analysis as a robustness check. 
Random aggregation still causes the same changes to the distribution of document lengths.
This allows us to see whether simply changing the document length increases the number of state-related topics to the same extent that we showcase in our study.
For the Twitter corpus we randomly assign legislators to tweets and rerun the legislator level 60 and 120-topic models. We complete this process ten times for a total of ten permutations. 
 Figure \ref{fig:permutations_Twitter} shows the distribution of state-related topics in the permuted models compared to the original legislator-level model. 
 
\begin{figure}[!htb]
    \centering
    \includegraphics[width=0.75\linewidth]{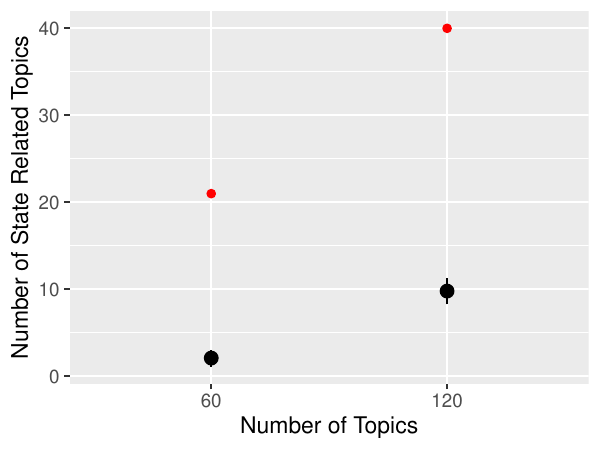}
    \caption{Distribution of state topics of the permuted birthplace models compared to original Wikipedia page models. Single red dots indicate the state-related topic counts of original models, black dots and whiskers indicate the mean state-related topic counts and the 95\% confidence interval around the permuted models.}
    \label{fig:permutations_wikipedia}
\end{figure}

We repeat the same exact procedure for the Wikipedia corpus, except this time assigning birthplaces randomly to articles.
Figure \ref{fig:permutations_wikipedia} shows these results.
 
In both instances, we find that the permuted models on average do not reach the number of state-related topic counts that the real non-permuted data does and that the actual counts are outside the 95\% confidence interval for the mean of the permuted model state-related topic counts.

 Our results suggest that the jump in the number of state-related topics when tweets are aggregated by legislators and when Wikipedia articles are aggregated by birthplaces is not just due to the changes to document length that result from aggregation.

\bibliographystyle{plainnat}
\bibliography{references}